\documentclass{article}
\usepackage{xcolor}
\usepackage{floatrow} 
\usepackage{caption}
\usepackage{subfigure}
\usepackage{wrapfig}
\usepackage{multirow}
\usepackage{authblk}
\usepackage{amsmath}
\usepackage{bm}
\usepackage{graphicx,floatrow}
\usepackage{algorithm}
\usepackage[noend]{algpseudocode}
\usepackage[final]{nips_2018}
\usepackage[T1]{fontenc}    
\usepackage{hyperref}       
\usepackage{url}            
\usepackage{booktabs}       
\usepackage{amsfonts}       
\usepackage{nicefrac}       
\usepackage{microtype}      
\usepackage{caption}
\setcitestyle{numbers,square}
\title{Knowledge Distillation by On-the-Fly Native Ensemble}

\author[1]{\textbf{Xu Lan}}
\author[2]{\textbf{Xiatian Zhu}}
\author[1]{\textbf{Shaogang Gong}}
\affil[1]{Queen Mary University of London}
\affil[2]{Vision Semantics Ltd}

\begin{document}

\maketitle

\begin{abstract}
	Knowledge distillation is effective to
	train small and generalisable network models
	for meeting the low-memory and fast running requirements.
	Existing offline distillation methods rely on a strong pre-trained teacher, 
	which enables favourable knowledge discovery and transfer but
	requires a complex two-phase training procedure.
	Online counterparts address this limitation at the price of
	lacking a high-capacity teacher.
	In this work, 
	we present an On-the-fly Native Ensemble (ONE) learning strategy 
	for one-stage online distillation. 
	%
	Specifically, ONE trains only a single multi-branch network 
	while simultaneously establishing a strong teacher on-the-fly
	to enhance the learning of target network.
	Extensive evaluations show that 
	ONE improves the generalisation performance a variety of deep neural networks
	more significantly than alternative methods
	on four image classification dataset: CIFAR10, CIFAR100, SVHN, and ImageNet,
	whilst having the computational efficiency advantages.
\end{abstract}

\section{Introduction}

Deep neural networks have gained impressive success
in many computer vision tasks 
\cite{krizhevsky2012imagenet,simonyan2014very,szegedy2015going,he2016deep,girshick2015fast,long2015fully,li2017person,lan2018person}.
However, the performance advantages are often gained 
at the cost of training and deploying resource-intensive 
networks with large depth and/or width \cite{zagoruyko2016wide,he2016deep,simonyan2014very}.
This leads to the necessity of developing compact yet still discriminative
models.
Knowledge distillation \cite{hinton2015distilling}
is one generic meta-solution among the others such as
parameter binarisation \cite{rastegari2016xnor,han2015deep} and
filter pruning \cite{li2016pruning}.
%
The distillation process begins with training 
a high-capacity {\em teacher} model (or an ensemble of networks), 
followed by learning a smaller {\em student} model 
which is encouraged to match the teacher's
predictions \cite{hinton2015distilling} and/or feature
representations \cite{ba2014deep,romero2014fitnets}.
Whilst promising the student model quality improvement
from aligning with a pre-trained teacher model,
this strategy requires a longer training time, significant extra computational cost 
and large memory (for heavy teacher)
with the need for a more complex multi-stage training process,
all of which are commercially unattractive \cite{anil2018large}.

To simplify the distillation training process as above, 
simultaneous  distillation
algorithms \cite{zhang2017deep,anil2018large} have been developed 
to perform online knowledge teaching in a one-phase learning
procedure.
Instead of pre-training a static teacher model, 
these methods train simultaneously a set of 
(typically two) student models which  
learn from each other in a peer-teaching manner.
This approach merges the training processes of 
the teacher and student models, and uses the peer network
to provide the teaching knowledge.
Beyond the original understanding of distillation
that requires the teacher model larger 
than the student, they allow to 
improve any-capacity model performance,
leading to a more generically applicable technique.
This peer-teaching strategy sometimes even 
outperforms the teacher based offline distillation,
with the plausible reason that the large teacher model
tends to overfit the training set and finally 
provides less information additional to the original training labels
\cite{anil2018large}.

However, existing online distillation has a number of drawbacks:
(1) Each peer-student may only provide limited extra information and 
resulting in suboptimal distillation;
(2) Training multiple students significantly increases the computational
cost and resource burdens;
(3) 
It requires asynchronous model updating
with a notorious need of carefully ordering the operations
of prediction and back-propagation across networks.  
We consider that all the weaknesses are due to the lacking of an appropriate teacher 
role in the online distillation processing.

In this work, we propose a novel online knowledge distillation method
that is not only more efficient (lower training cost) 
and but also more effective (higher model generalisation improvement) 
as compared to previous alternative methods.
In {\em training}, the proposed approach 
constructs a multi-branch variant of a given target network
by adding auxiliary branches,
creates a native ensemble teacher model from all branches on-the-fly,
and learns simultaneously each branch plus the teacher model
subject to the same target label constraints.
Each branch is trained with two objective loss terms:
a conventional softmax cross-entropy loss
which matches with the ground-truth label distributions,
and a distillation loss which 
aligns to the teacher's prediction distributions. 
Comparing with creating a set of student networks,
a multi-branch single model is more efficient to train 
whilst achieving superior generalisation performance and 
avoiding asynchronous model update.  
In {\em test}, we simply convert the trained multi-branch model back to 
the original (single-branch) network architecture by removing
the auxiliary branches,
therefore introducing no test-time cost increase.
In doing so, we derive an {\bf On-the-Fly Native Ensemble} (ONE) teacher 
based simultaneous
distillation training approach that not only eliminates 
the computationally expensive need for pre-training the teacher model
in an isolated stage as the offline counterparts,
but also further improves the quality of online distillation.

Extensive comparative experiments on four benchmarks (CIFAR10/100, SVHN, and ImageNet) 
show that the proposed ONE distillation method
enables to train more generalisable target models in a one-phase
process than the alternative strategies of
offline learning a larger teacher network or
simultaneously distilling peer students,
the previous state-of-the-art techniques
for training small target models.

\section{Related Work}

\textbf{Knowledge Distillation.} 
There have been a number of attempts to
transfer knowledge between varying-capacity network models \cite{bucilua2006model,hinton2015distilling,ba2014deep,romero2014fitnets}. 
Hinton et al. \cite{hinton2015distilling} distilled knowledge 
from a large pre-trained teacher model to improve a small target net.
The rationale behind is the use of extra supervision from teacher model in target model training, beyond a conventional supervised learning objective such as the cross-entropy loss subject to labelled training data. The extra supervision were typically obtained from a pre-trained powerful teacher model
in the form of classification probabilities \cite{hinton2015distilling},
feature representation \cite{ba2014deep,romero2014fitnets},
or inter-layer flow (the inner product of feature maps) \cite{yim2017gift}.

Recently, knowledge distillation has been exploited to distil easy-to-train large networks 
into harder-to-train small networks \cite{romero2014fitnets}.
Previous distillation methods often take offline learning strategies,
requiring at least two phases of training.
The more recently proposed deep mutual learning \cite{zhang2017deep}
overcomes this limitation by conducting online distillation in one-phase training between two peer student models.
Anil et al. \cite{anil2018large} further extended this idea 
to accelerate large scale distributed neural network training.
%

However, existing online distillation methods 
lacks a strong  ``teacher'' model which limits the efficacy of knowledge discovery and
transfer. Like offline counterpart, multiple nets are needed to be trained and therefore
computationally expensive. 
We overcome both limitations by designing 
a new variant of online distillation training algorithm
characterised by simultaneously learning a teacher on-the-fly
and the target net and performing batch-wise knowledge transfer
in a one-phase procedure.


\vspace{0.1cm}
\noindent \textbf{Multi-branch Architectures.} 
Multi-branch based neural networks have been widely exploited in computer vision tasks \cite{szegedy2015going,szegedy2016rethinking,he2016deep}. 
For example, ResNet \cite{he2016deep} can be thought of 
as a category of two-branch networks
where one branch is the identity mapping.  
Recently, 
``grouped convolution'' \cite{xie2017aggregated,huang2017condensenet} 
has been used as a replacement of standard convolution
in constructing  multi-branch net architectures.
These building blocks are often utilised as templates
to build deeper networks for achieving stronger modelling capacity.
Whilst sharing the multi-branch principle,
our ONE method is fundamentally different from these above existing methods
since our objective is to improve the training quality of
any given target network,
rather than presenting a new multi-branch building block.
In other words, our method is a meta network learning algorithm,
independent of specific network architectures.



\section{Knowledge Distillation by On-the-Fly Native Ensemble}

\begin{figure*} [h]
	\centering
	\includegraphics[width=1.0\linewidth]{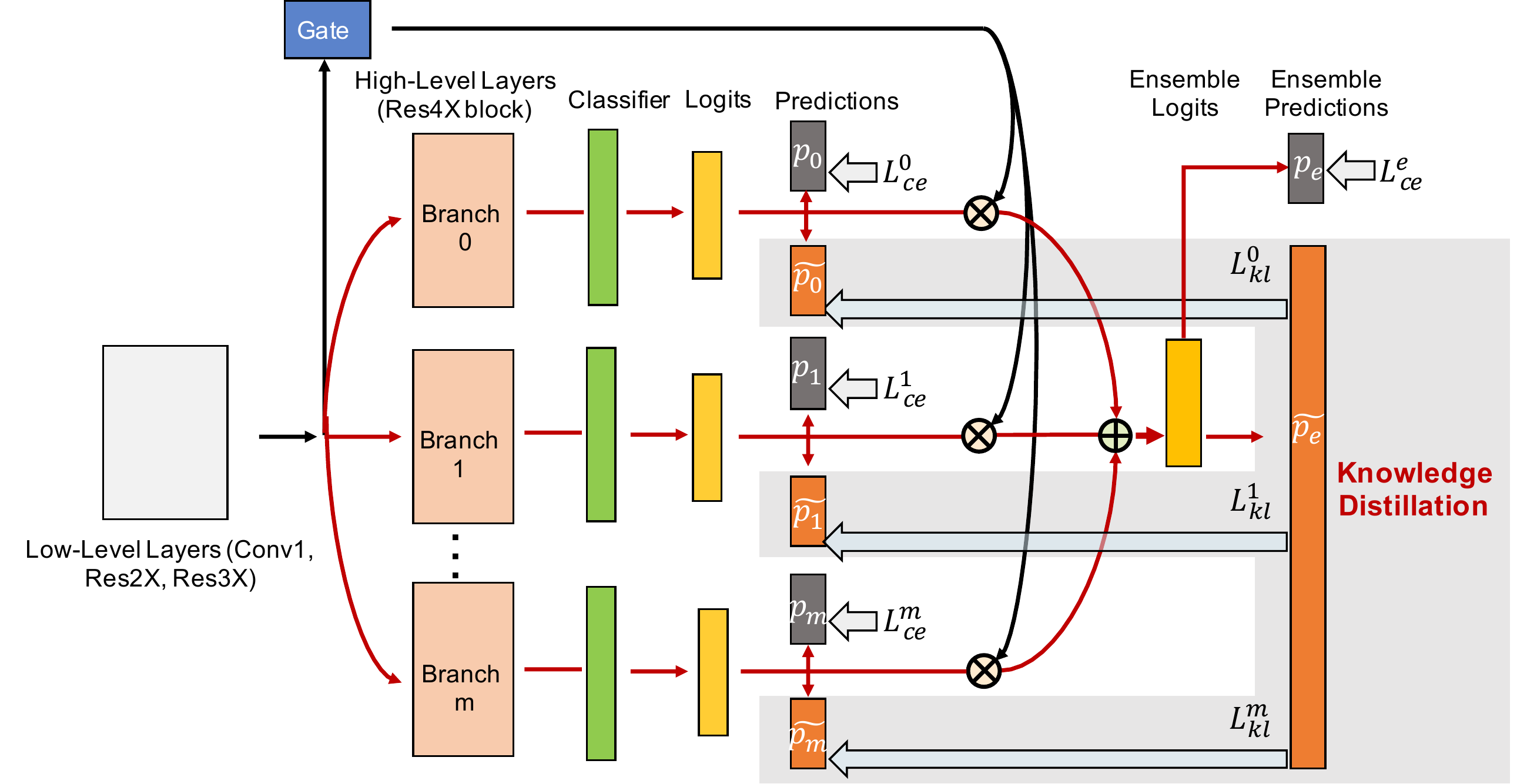} 	
	\caption{
		Overview of online distillation training of ResNet-110 
		by the proposed On-the-fly Native Ensemble (ONE).
		With ONE, we reconfigure the network by
		adding $m$ auxiliary branches which share
		the low-level layers with the target net. 
		Each branch with shared layers makes an individual model, 
		and their ensemble is used to build the teacher model.
		During a mini-batch training process, 
		we employ the teacher to collect knowledge from individual
		branch models on-the-fly, which in turn is distilled back to all branches
		to enhance model learning in a close-loop form.
		In test, auxiliary branches can be either discarded
		or kept based on the deployment efficiency requirement.}
	\label{fig:architecture}
\end{figure*}

We formulate an online distillation training method 
based on a concept of On-the-fly Native Ensemble (ONE).
For understanding convenience, we take ResNet-110 \cite{he2016deep} on CIFAR100 dataset as an example.
It is straightforward to apply ONE to other network architectures.
For model training,
we often have access to $n$ labelled training samples 
$\mathcal{D}=\{(\bm{x}_{i}, {y}_{i})\}_{i}^{n}$ with  
each belonging to one of $C$ classes 
${y}_{i} \in \mathcal{Y} = \{1,2,\cdots,C \}$.
The network $\bm{\theta}$ outputs a probabilistic class posterior  
$p(c | \bm{x}, \bm{\theta} )$ 
for a sample $\bm{x}$ over a class $c$ as: 
\begin{equation}
\label{eq:softmax}
p(c | \bm{x}, \bm{\theta} ) = {f}_{sm}(\bm z)=
\frac{\exp(\bm z^c)} {\sum_{j=1}^{C} \exp(\bm z^j)}, \;\; 
c \in \mathcal{Y}
\end{equation}
where $\bm{z}$ is the logits or unnormalised log probability
outputted by the network $\bm{\theta}$.
To train a multi-class classification model, 
we often adopt the Cross-Entropy (CE) measurement between
the predicted and ground-truth label distributions as the objective function:
\begin{equation}
\mathcal{L}_\text{ce} = - 
\sum_{c=1}^C \delta_{c,y} \log \Big(p({c}|\bm{x}, \bm{\theta}) \Big)
\label{eq:cross_loss}
\end{equation}
where $\delta_{c,y}$ is Dirac delta which returns 1 if $c$ is the ground-truth label, and 0 otherwise.
With the CE loss, the network is trained to 
predict the correct class label 
in a principle of maximum likelihood.
To further enhance the model generalisation,
we concurrently distil extra knowledge 
from an on-the-fly native ensemble (ONE) teacher in training.

\vspace{0.1cm}
\noindent {\bf On-the-Fly Native Ensemble.}
Overview of the ONE architecture is depicted in Fig \ref{fig:architecture}.
The ONE consists of two components:
(1) $m$ auxiliary branches with the same configuration
(Res4X block and an individual classifier), 
each of which serves as an independent classification model
with shared low-level stages/layers.
This is because low-level features are largely shared 
across different network instances which allows to
reduce the training cost.   
(2) A gate component which learn to
ensemble all $(m\!+\!1)$ branches to build a stronger 
teacher model. 
It is constructed by one FC layer followed by 
batch normalisation, ReLU activation, and softmax,
and uses the same input features as the branches.

Our ONE method is established based on a multi-branch design specially for
model training with several merits:
(1) Enable the possibility of creating a strong teacher model
without training a set of networks at a high computational cost;
(2) Introduce a multi-branch simultaneous learning regularisation 
which benefits model generalisation (Fig \ref{Fig:model_component_b});
(3) Avoid the tedious need for asynchronous update between
multiple networks.

Under the reconfiguration of network,
we add a separate CE loss $\mathcal{L}_\text{ce}^i$ to all branches 
which simultaneously learn to predict the 
same ground-truth class label of a training sample.
While sharing the most layers, each branch
can be considered as an independent multi-class classifier
in that all of them independently learn high-level semantic representations.
Consequently, taking the ensemble of all branches (classifiers)
can make a stronger teacher model.
One common way of ensembling models is 
to average individual predictions.
This may ignore the diversity and importance variety 
of member models in the ensemble.
We therefore learn to ensemble by the gating component as:
\begin{equation}\label{eq:Gate_combine}
\bm z_e = \sum\limits_{i=0}^{m} g_i \cdot \bm z_i
\end{equation}
where $g_i$ is the importance score of the $i$-th branch's logits $\bm z_i$,
and $\bm z_e$ is the logits of the ONE teacher.
In particular, we denote the original branch as $i=0$
for indexing convenience.
We train the ONE teacher model with the CE loss $\mathcal{L}_\text{ce}^e$ (Eq \eqref{eq:cross_loss})
same as the branch models.

\vspace{0.1cm}
\noindent {\bf Knowledge Distillation.}
Given the teacher logits for each training sample, 
we distil knowledge back into all branches in a closed-loop form.
For facilitating knowledge transfer,
we compute soft probability distributions at a temperature of $T$
for individual branches and the ONE teacher as:
\begin{equation}
\label{eq:softmax_soft_S}
\tilde{p}_i(c | \bm{x}, \bm{\theta^}{i}) = 
\frac{\exp(\bm z_i^c/T)} {\sum_{j=1}^{C} \exp(\bm z_i^j/T)}, 
c \in \mathcal{Y}
\end{equation}
\begin{equation}
\tilde{p}_e(c | \bm{x}, \bm{\theta^}{e}) = \frac{\exp(\bm z_e^c/T)} {\sum_{j=1}^{C} \exp(\bm z_e^j/T)},
c \in \mathcal{Y}
\end{equation}
where $i$ denotes the branch index, $i\!=\!0,\cdots,m$,
$\bm{\theta^}{i}$ and $\bm{\theta^}{e}$
the parameters of branch and teacher models respectively.
Higher values of $T$ lead to more softened distributions.

To quantify the alignment between individual branches and the teacher in
their predictions, we use the Kullback Leibler divergence written as:
\begin{equation} 
\label{eq:kt_loss}
\mathcal{L}_\text{kl}= \sum_{i=0}^{m}  \sum_{j=1}^{C}   
{ \tilde{p}_e(j|\bm{x}, \bm{\theta}^e)} \log \frac {\tilde{p}_e(j|\bm{x}, \bm{\theta}^e)}{{\tilde{p}_i(j|\bm{x}, \bm{\theta}^i)}}.
\end{equation}

\vspace{0.1cm}
\noindent{\bf Overall Loss Function.}
We obtain the overall loss function for online distillation training 
by the proposed ONE as:
\begin{equation} 
\label{eq:Total_loss}
\mathcal{L}= \sum_{i=0}^{m} \mathcal{L}_\text{ce}^i +
\mathcal{L}_\text{ce}^e + T^2 * \mathcal{L}_\text{kl}
\end{equation}
where $\mathcal{L}_\text{ce}^i$ and $\mathcal{L}_\text{ce}^e$ 
are the conventional CE loss terms
associated with the $i$-th branch and the ONE teacher, respectively.
The gradient magnitudes produced by the soft targets $\tilde{p}$ 
are scaled by $\frac{1}{T^2}$, 
so we multiply the distillation loss term by
a factor ${T^2}$ to ensure that 
the relative contributions of ground-truth and teacher probability distributions 
remain roughly unchanged.
%
Following \cite{hinton2015distilling}, we set $T=3$ in our all experiments.

\begin{algorithm}
	\caption{Knowledge Distillation by On-the-Fly Native Ensemble}
	\label{alg}
	\begin{algorithmic}[1]
		\State{{\bf Input}: 
			Labelled training data $\mathcal{D}$; 
			Training epochs $\tau$;
			Auxiliary branch number $m$;
		}
		\State{{\bf Output}: 
			Trained target CNN model $\bm{\theta}^0$,
			and auxiliary models $\{ \bm{\theta}^i \}_{i=1}^m$;
		}
		
		\vspace{0.1cm}
		\State{{\bf /* Training */} }
		\State{{\bf Initialisation}: 
			t=1; 
			Randomly initialise $\{ \bm{\theta}^i \}_{i=0}^m$;
		}
		
		\While{$t \leq \tau$}
		
		\State Compute predictions 
		of all individual branches $\{ p_i \}_{i=0}^m$ (Eq \eqref{eq:softmax});
		
		\State Compute the teacher logits (Eq \eqref{eq:Gate_combine});
		
		\State Compute the soft targets of all branches and teacher (Eq \eqref{eq:softmax_soft_S});
		
		\State Distil knowledge from the teacher to all branches (Eq \eqref{eq:kt_loss});
		
		\State Obtain the final loss function (Eq \eqref{eq:Total_loss});
		
		\State Update the model parameters $\{ \bm{\theta}^i \}_{i=0}^m$ by SGD.

		
		\EndWhile
		
		\State \textbf{end}
		
		\vspace{0.1cm}
		\State{{\bf /* Testing */} }
		\State         
		{\bf Single model deployment:} Use $\bm{\theta}^0$; 
		\State   
		{\bf Ensemble deployment (ONE-E):} Use $\{ \bm{\theta}^i \}_{i=0}^m$.     
	\end{algorithmic}
\end{algorithm}

\vspace{0.1cm}
\noindent{\bf Model Training and Deployment.}
The model optimisation and deployment details are summarised in Alg \ref{alg}.
Unlike the two-phase offline distillation training,
the target network and the ONE teacher are trained simultaneously
and collaboratively, with the knowledge distillation 
from the teacher to the target
being conducted in each mini-batch 
and throughout the whole training procedure. 
Since there is only one multi-branch network 
rather than multiple networks,
we only need to carry out the same stochastic gradient descent 
through $(m+1)$ branches, and training the whole network until convergence, 
as the standard single-model incremental batch-wise training.
There is no complexity of asynchronous updating
among different networks which is required in deep mutual learning \cite{zhang2017deep}.

Once the model is trained, we simply remove 
all the auxiliary branches and obtain the original 
network architecture for deployment.
Hence, our ONE method does not increase
test-time cost.
However, if there is less constraint on computation budget
and model performance is more important,
we can deploy it as an ensemble with all trained branches,
denoted as {``ONE-E''}.

\section{Experiments}

{\bf Datasets.}
We used four multi-class categorisation benchmark datasets in our evaluations. 
\textbf{(1)} {\em {CIFAR-10}} \cite{krizhevsky2009learning}:
A natural images dataset that contains
50,000/10,000 training/test samples drawn from 10 object classes (in total 60,000 images). 
Each class has 6,000 images sized at   $32\!\times\!32 $  pixels.
%
\textbf{(2)} {\em CIFAR-100} \cite{krizhevsky2009learning}: 
A similar dataset as CIFAR10 that also contains 50,000/10,000
training/test images but covering 100 fine-grained classes. 
Each class has 600 images.
\textbf{(3)} {\em SVHN}: The Street View House Numbers (SVHN) dataset consists of 73,257/26,032 standard training/text images and an extra set of 531,131 training images. 
We used all the training data without data augmentation
as \cite{huang2016densely,lee2015deeply}.
\textbf{(4)} {\em ImageNet}: 
The 1,000-class dataset from ILSVRC 2012  \cite{russakovsky2015imagenet}
provides 1.2 million images for training, 
and 50,000 for validation.

\vspace{0.1cm}
\noindent{\bf Performance metric.}
We adopted the common top-$n$ ($n$=1, 5) classification error rate. 
For computational cost of model training and test, 
we used the criterion of floating point operations (FLOPs).
For any network trained by ONE, we report the average
performance of all branch outputs with standard deviation.

\vspace{0.1cm}
\noindent{\bf Experiments setup.}
We implemented all networks and model training procedures in Pytorch.
For all datasets, 
we adopted the same experimental settings as \cite{huang2016deep,xie2017aggregated} for making fair comparisons.
We used the SGD with Nesterov momentum  and set the momentum to 0.9, following a standard learning rate schedule that drops 
from 0.1 to 0.01 at 50\% training 
and to 0.001 at 75\%. 
For the training budget, 
CIFAR/SVHN/ImageNet used 300/40/90 epochs respectively.
We adopt a 3-branch ONE ($m\!=\!2$) design unless stated otherwise. 
We separate the last block 
of each backbone net from the parameter sharing
(except on ImageNet we separate the last 2 blocks 
for giving more learning capacity to branches) without extra structural optimisation  
(see ResNet-110 for example in Fig \ref{fig:architecture}).
Most state-of-the-art nets are in block structure designs.

\begin{table*}
	\setlength{\tabcolsep}{0.2cm} 
	\caption{
		Evaluation of our ONE method on CIFAR and SVHN.
		Metric: Error rate (\%). 
	}
	\label{Single_branch}
	\centering
	\begin{tabular}{l|l|l|l||c}
		\hline 
		Method     & CIFAR10 & CIFAR100 & SVHN & Params\\ 
		\hline
		\hline
		ResNet-32 \cite{he2016deep}      & 6.93  &31.18 &2.11&0.5M   \\
		ResNet-32 + {\bf ONE} &\bf5.99$\pm$0.05 &\bf26.61$\pm$0.06 &\bf1.83$\pm$0.05 &0.5M\\
		\hline
		ResNet-110 \cite{he2016deep}     &5.56  &25.33 &2.00 &1.7M  \\
		ResNet-110 + {\bf ONE}    &\bf5.17$\pm$0.07  &\bf 21.62$\pm$0.26 &\bf1.76$\pm$0.07 &1.7M   \\
		\hline
		ResNeXt-29($8\!\times\!64d$) \cite{xie2017aggregated}
		& 3.69 & 17.77 & 1.83 &34.4M\\
		ResNeXt-29($8\!\times\!64d$) + {\bf ONE} 
		&\bf3.45$\pm$0.04&\bf16.07$\pm$0.08  & \bf1.70$\pm$0.03&34.4M\\
		\hline 
		DenseNet-BC(L=190, k=40) \cite{huang2017densely}&	3.32&17.53&1.73&25.6M\\
		DenseNet-BC(L=190, k=40) + {\bf ONE}&\bf3.13$\pm$0.07	&\bf16.35$\pm$0.05&\bf1.63$\pm$0.05&25.6M\\	 
		\hline 
	\end{tabular}
\end{table*}

\subsection{Evaluation of On-the-Fly Native Ensemble}

{\bf Results on CIFAR and SVHN. }
Table \ref{Single_branch} compares top-1 error rate performances
of four varying-capacity state-of-the-art network models trained by
the conventional and our ONE learning algorithms.
We have these observations:
(1) All different networks benefit from the ONE training algorithm,
particularly for small models achieving larger performance gains.
This suggests the generic superiority of our method
for online knowledge distillation from 
the on-the-fly teacher to the target model.
(2) All individual branches have similar performances,
indicating that they have made sufficient agreement and 
exchanged respective knowledge to each other well through the
proposed ONE teacher model during training.

\begin{table}[h]
	\centering 
	\caption{
		Evaluation of our ONE method on ImageNet.
		Metric: Error rate (\%). 
	}
	\setlength{\tabcolsep}{0.1cm}
	\label{Imagenet}
	\centering
	\begin{tabular}{l|l|l}
		\hline
		Method  & Top-1 & Top-5 \\ \hline \hline
		ResNet-18 \cite{he2016deep} 
		& 30.48 & 10.98  \\
		ResNet-18 + {\bf ONE}   
		
		&\bf 29.45$\pm$0.23 & \bf 10.41$\pm$0.12 \\
		\hline \hline
		ResNeXt-50 \cite{xie2017aggregated} 
		& 22.62 &6.29 \\
		ResNeXt-50 + {\bf ONE}  
		&\bf21.85$\pm$0.07&\bf5.90$\pm$0.05  \\
		\hline \hline
		SeNet-ResNet-18 \cite{hu2017squeeze} 
		& 29.85 &10.72 \\
		SeNet-ResNet-18 + {\bf ONE}  
		&\bf29.02$\pm$0.17&\bf10.13$\pm$0.12  \\
		\hline %
	\end{tabular}
\end{table}

\vspace{0.2cm}
\noindent{\bf Results on ImageNet. }
Table \ref{Imagenet} shows the comparative performances on the 1000-classes ImageNet. 
It is shown that the proposed ONE learning algorithm still
yields more effective training and more generalisable models
in comparison to vanilla SGD.
This indicates that our method can be generically applied to large scale 
image classification settings.

\begin{table*}
	\centering 
	\caption{
		Comparison with knowledge distillation methods
		on CIFAR100.
		``*'': Reported results.
		TrCost/TeCost: Training/test cost, in unit of 10$^{8}$ FLOPs.
		{\bf {\color{red}Red}/{\color{blue}Blue}}:
		Best and second best results.
	}
	\setlength{\tabcolsep}{0.25cm}
	\label{tab:KD}
	\centering
	\begin{tabular}{l|l|c|c||l|c|c}
		\hline
		Target Network 
		& \multicolumn{3}{c||}{ResNet-32}
		& \multicolumn{3}{c}{ResNet-110} 
		\\ \hline 
		Metric & Error (\%) & TrCost & TeCost 
		& Error (\%) & TrCost & TeCost \\
		\hline \hline
		KD \cite{hinton2015distilling}
		&\bf\color{blue}28.83 & 6.43& 1.38
		& N/A & N/A & N/A 
		\\
		DML \cite{zhang2017deep}
		& 29.03$\pm$0.22$^*$ &\bf\color{blue} 2.76& 1.38
		& \bf\color{blue} 24.10$\pm$0.72 &\bf\color{blue}10.10&5.05
		\\
		\hline
		\bf ONE
		& \bf\color{red}26.61$\pm$0.06 &\bf\color{red} 2.28 &  1.38
		& \bf\color{red}21.62$\pm$0.26 &\bf\color{red} 8.29 &5.05
		\\ \hline
	\end{tabular}
\end{table*}

\subsection{Comparison with Distillation Methods}

We compared our ONE method with two representative distillation methods:
Knowledge Distillation (KD) \cite{hinton2015distilling} and
Deep Mutual Learning (DML) \cite{zhang2017deep}.
For the offline competitor KD,
we used a large network ResNet-110 as the teacher  
and a small network ResNet-32 as the student.
For the online methods DML and ONE, we evaluated their performance
using either ResNet-32 or ResNet-110 as the target model.
We observed from Table \ref{tab:KD} that: 
(1) ONE outperforms both KD (offline) and DML (online) distillation
methods in error rates,
validating the performance advantages of our method
over alternative algorithms when applied to different
CNN models.
(2) ONE takes the least model training cost
and the same test cost as others,
and therefore leading to the most cost-effective solution.

\begin{table*}
	\centering 
	\caption{
		Comparison with ensembling methods
		on CIFAR100.
		``*'': Reported results.	
		TrCost/TeCost: Training/test cost, in unit of 10$^{8}$ FLOPs.
		{\bf {\color{red}Red}/{\color{blue}Blue}}:
		Best and second best results.
	}
	\setlength{\tabcolsep}{0.3cm}
	\label{tab:ensemble}
	\centering
	\begin{tabular}{l|c|c|c||c|c|c}
		\hline
		Network 
		& \multicolumn{3}{c||}{ResNet-32}
		& \multicolumn{3}{c}{ResNet-110} 
		\\ \hline 
		Metric & Error (\%) & TrCost & TeCost 
		& Error (\%) & TrCost & TeCost \\
		\hline \hline
		Snopshot Ensemble \cite{huang2017snapshot}
		& 27.12 &\color{red}\bf 1.38 & 6.90
		&  23.09$^*$\ &\bf\color{red} 5.05&25.25
		\\
		2-Net Ensemble
		& 26.75 & 2.76&\bf\color{blue}  2.76
		& 22.47 & 10.10 &\bf\color{blue} 10.10
		\\
		3-Net Ensemble
		& \bf\color{blue}25.14 & 4.14& 4.14
		&\bf\color{blue} 21.25&15.15&15.15
		\\ \hline
		\bf ONE-E
		& \bf \color{red} 24.63 &\bf\color{blue} 2.28 &\color{red} \bf 2.28
		& \bf \color{red} 21.03 &\bf\color{blue} 8.29 &\color{red} \bf 8.29
		\\
		\hline \hline
		{\bf ONE} &
		26.61 & 2.28&  1.38
		&21.62&8.29&5.05
		\\
		\hline
	\end{tabular}
	
\end{table*}

\subsection{Comparison with Ensembling Methods}

Table \ref{tab:ensemble} compares the performances 
of our multi-branch (3 branches) based model ONE-E and standard ensembling methods. 
It is shown that ONE-E yields not only the best test error 
but also allows for most efficient deployment with
lowest test cost. 
These advantages are achieved at 
second lowest training cost.
Whilst Snapshot Ensemble takes the least training cost,
its generalisation performance is unsatisfied 
with a notorious drawback of highest deployment cost.

It is worth noting that ONE (without branch ensemble) already outperforms comprehensively
2-Net Ensemble in terms of error rate, training and test cost. 
Comparing 3-Net Ensemble, ONE is able to approach the generalisation performance
whilst having even larger model training and test efficiency advantages.

\begin{table*} [h]
	\centering \caption{Model component analysis on CIFAR100.
		Network: ResNet-110.
	}
	\setlength{\tabcolsep}{0.25cm}
	\label{tab:Model_components}
	\centering
	\begin{tabular}{l|c|c|c|c}
		\hline
		Configuration 
		& Full 
		& W/O Online Distillation 
		& W/O Sharing Layers
		& W/O Gating \\ \hline \hline
		ONE
		&\bf 21.62$\pm$0.26
		& 24.73$\pm$0.20 
		& 22.45$\pm$0.52
		& 22.26$\pm$0.23
		\\ \hline
		ONE-E
		& 21.03
		& 21.84 
		& \bf 20.57 
		& 21.79
		\\ \hline
	\end{tabular}
\end{table*}

\subsection{Model Component Analysis}

\begin{figure}[htbp]\centering                                                          
	\subfigure[ONE without online distillation]{                    
		\begin{minipage}{7cm}\centering  \label{Fig:model_component_b}                                                          
			\includegraphics[scale=0.45]{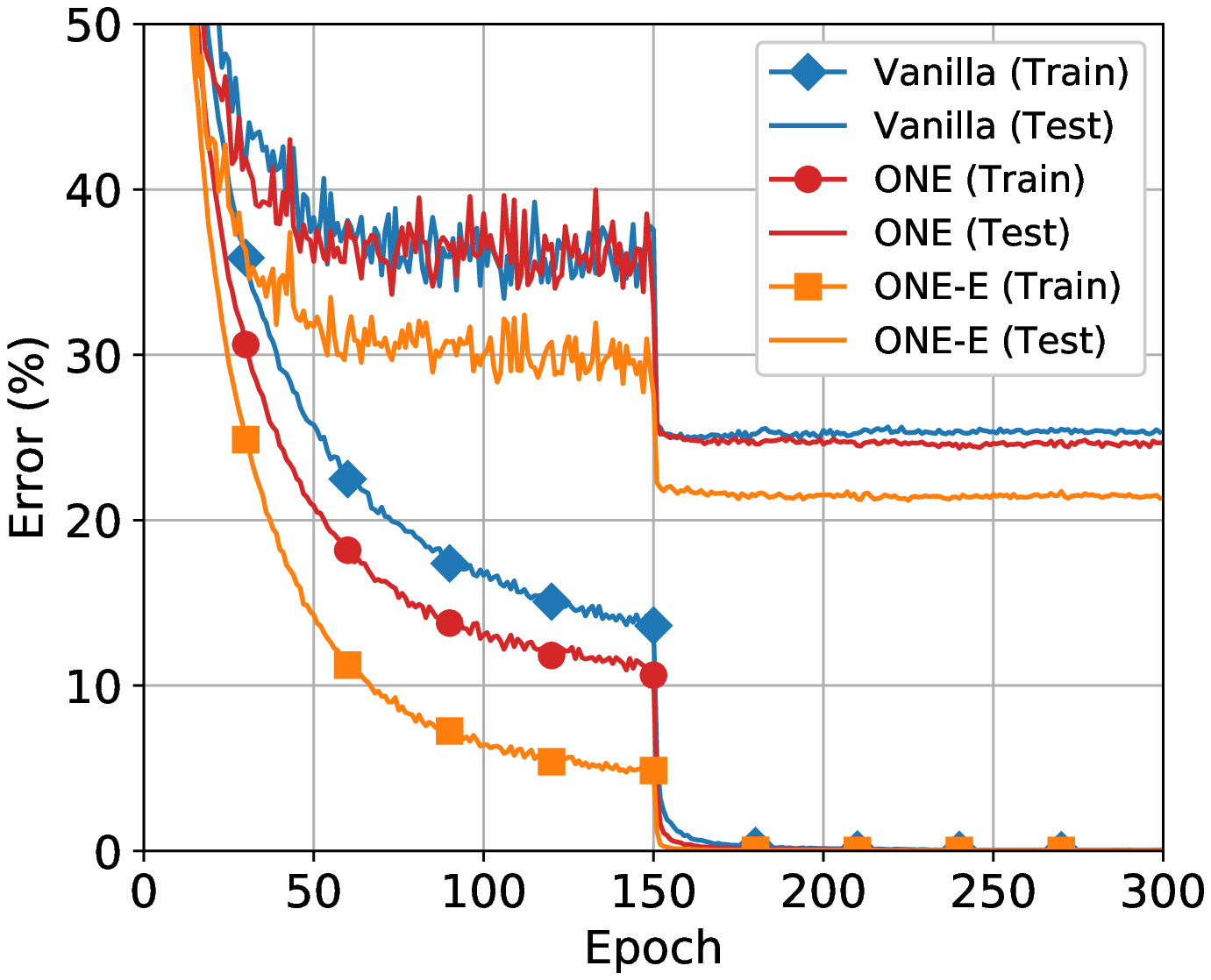}               
	\end{minipage}}\subfigure[Full ONE model]{                   
		\begin{minipage}{7cm}\centering \label{Fig:model_component_c}                                                             
			\includegraphics[scale=0.45]{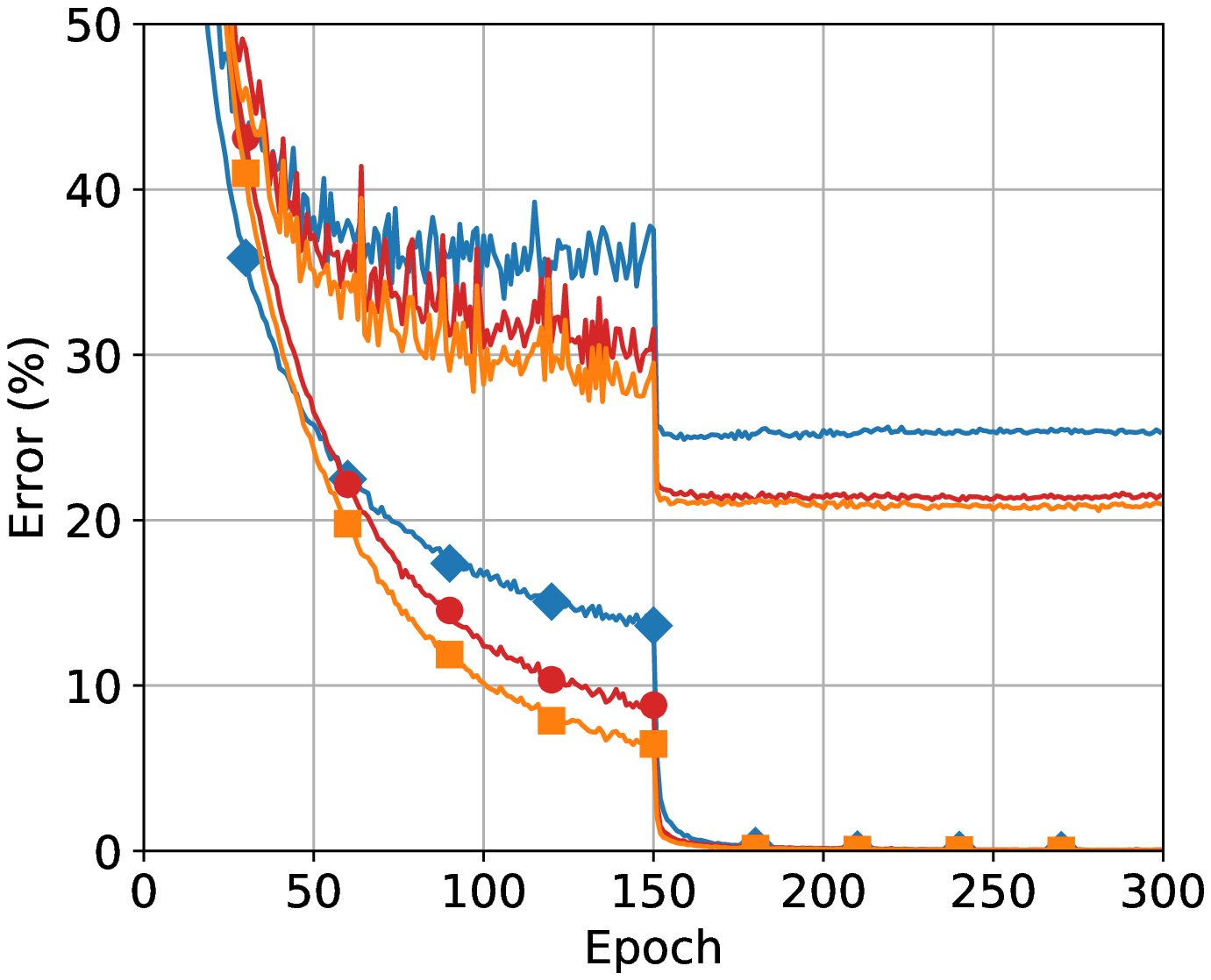}                
	\end{minipage}}\caption{
		Effect of online distillation.
		Network: ResNet-110.} 
	\label{Fig:model_component}  
	\vspace{-0.3cm}   
\end{figure}

Table \ref{tab:Model_components} shows the benefits of individual ONE components
on CIFAR100 using ResNet-110.
We have these observations:
(1) 
{\bf Without online distillation} 
(Eq \eqref{eq:kt_loss}), 
the target network suffers a performance drop of $3.11\%$ ($24.73$-$21.62$) in test error rate.
This performance drop validates the efficacy and quality of ONE teacher
in terms of performance superiority over individual branch models.
This can be more clearly seen in Fig \ref{Fig:model_component} that ONE teacher fits better to training data
and generalises better to test data.
Due to the closed-loop design,
ONE teacher also mutually benefits from distillation,
reducing its error rate from $21.84\%$ to $21.03\%$.
With distillation, the target model effectively
approaches ONE teacher (Fig \ref{Fig:model_component_b} vs. \ref{Fig:model_component_c})
on both training and test error performance,
indicating the success of teacher knowledge transfer.
Interestingly, even without distillation,
ONE still achieves 
better generalisation than the vanilla algorithm.
This suggests that our multi-branch design brings some positive regularisation effect
by concurrently and jointly learning the shared low-level layers
subject to more diverse high-level representation knowledge.
(2) {\bf Without sharing the low-level layers} not only increases
the training cost ($83\%$ increase), but also leads to 
weaker performance ($0.83\%$ error rate increase).
The plausible reason is the lacking of multi-branch regularisation effect
as indicated in Fig \ref{Fig:model_component_b}.
(3) Using average ensemble of branches {\bf without gating} (Eq \eqref{eq:Gate_combine})
causes a performance decrease of $0.64\%$($22.26$-$21.62$).
This suggests the benefit of adaptively exploiting the branch diversity in forming the ONE teacher.

\begin{table} 
	\centering \caption{
		Benefit of adding branches to ONE on CIFAR100.
		Network: ResNet-32.
	}
	\setlength{\tabcolsep}{0.3cm}
	\label{tab:Model_branch}
	\centering
	\begin{tabular}{l|c|c|c|c|c}
		\hline 
		Branch \#  & 1 & 2 & 3 & 4 & 5 \\ \hline \hline
		Error (\%) 
		& 31.18
		&27.38
		&26.68
		&26.58
		&\bf 26.52 \\
		\hline 
	\end{tabular}
\end{table}

The main experiments use 3 branches in ONE. 
Table \ref{tab:Model_branch} shows that 
ONE scales well with more branches
and the ResNet-32 model generalisation improves on CIFAR100 with 
the number of branches added during training
hence its performance advantage over the independently trained network
($31.18\%$ error rate).

\subsection{Model Generalisation Analysis}
We aim to give insights on why 
ONE trained networks yield 
a better generalisation capability.
A few previous studies \cite{keskar2016large,chaudhari2016entropy}
demonstrate that
the width of a local optimum is related to model generalisation.
The general understanding is that, 
the surfaces of training and test error
largely mirror each other and it is favourable to converge the models 
to broader optima.
As such, the trained model remains approximately optimal even under small perturbations
in test time.
Next, we exploited this criterion to examine the quality of model solutions
$\bm \theta_v$, $\bm \theta_m$, $\bm \theta_o$
discovered by the vanilla, DML and ONE training algorithms respectively.
This analysis was conducted on CIFAR100 using ResNet-110.

\begin{figure}[htbp]\centering                                                       
	\subfigure[Robustness on training data]{                   
		\begin{minipage}{7cm}\centering \label{Fig:localminima_loss}                                                         
			
			\includegraphics[scale=0.33]{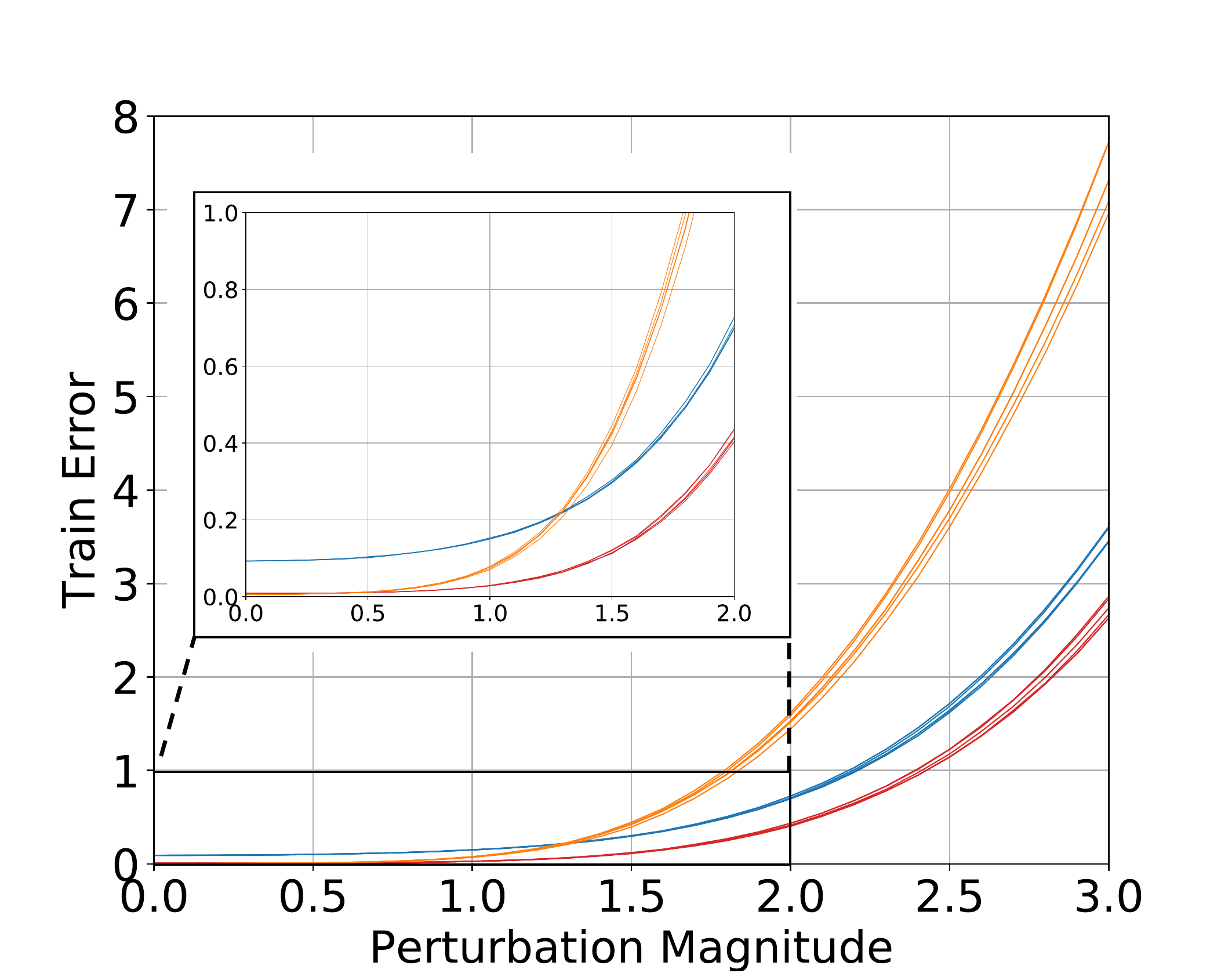}                
	\end{minipage}}\subfigure[Robustness on test data]{                    
		\begin{minipage}{7cm}\centering  
			\label{Fig:localminima_error}                                                        
			\includegraphics[scale=0.33]{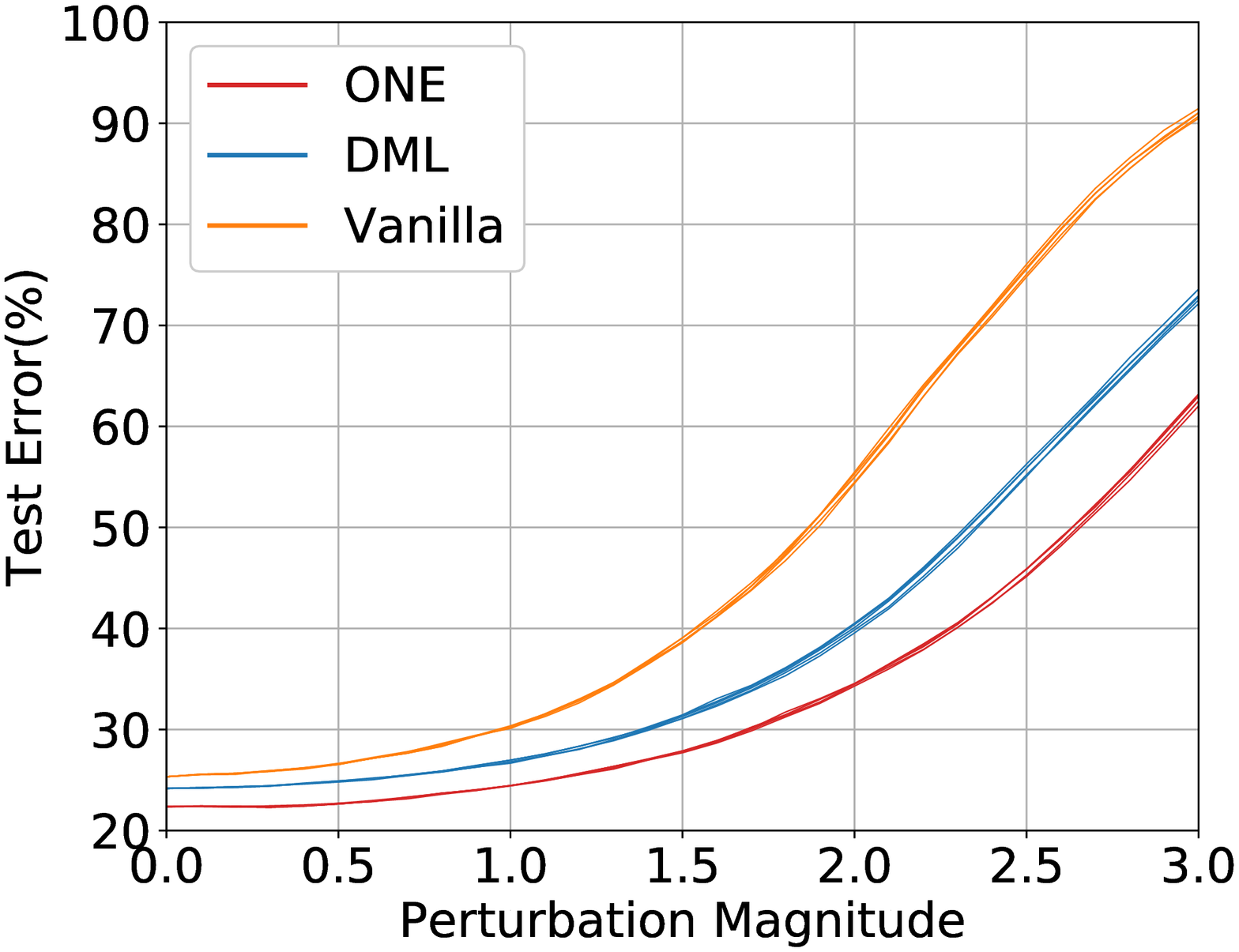}               
	\end{minipage}}\caption{
		Robustness test of ResNet-110 solutions found by ONE, DML, and vanilla
		training algorithms on CIFAR100.
		Each curve corresponds to a specific perturbation direction $\bm{v}$. 
	}   
	\label{Fig:localminima} 
	\vspace{-0.3cm}                                                            
\end{figure} 
Specifically, to test the width of local optimum, 
we added small perturbations to the solutions as
$\bm{\theta}_*(d, \bm v) \! = \! \bm \theta_* \! + \! d \cdot \bm v, \; 
* \! \in \! \{v,m,o\}$
where $\bm v$ is a uniform distributed direction vector with a unit length,
and $d \! \in \! [0,5]$ controls the change magnitude.
At each magnitude scale, we further sampled randomly 5 different direction vectors
to disturb the solutions.
We then tested the robustness of all perturbed models
in training and test error rates.
The training error is quantified as the cross-entropy measurement
between the predicted and ground-truth label distributions.

We observed in Fig \ref{Fig:localminima} that:
(1) The robustness of each solution against 
parameter perturbation appears to indicate
the width of local optima as:
$\bm \theta_v \! < \! \bm \theta_m \! < \! \bm \theta_o$.
{That is, ONE seems to find the widest local minimum among three 
	therefore more likely to generalise better than others.}

(2) Comparing with DML, vanilla and ONE found deeper local optima with lower training errors.
This indicates that DML may probably get stuck in training, 
therefore scarifying the vanilla's exploring capability
for more generalisable solutions to exchange 
the ability of identifying wider optima.
In contrast, our method further improves
the capability of identifying wider minima over DML whilst
maintaining the original exploration quality.
(3) For each solution, the performance changes on training and test data
are highly consistent, confirming the earlier 
observation \cite{keskar2016large,chaudhari2016entropy}.

\begin{wrapfigure}{r}{6cm}
	\includegraphics[width=6cm]{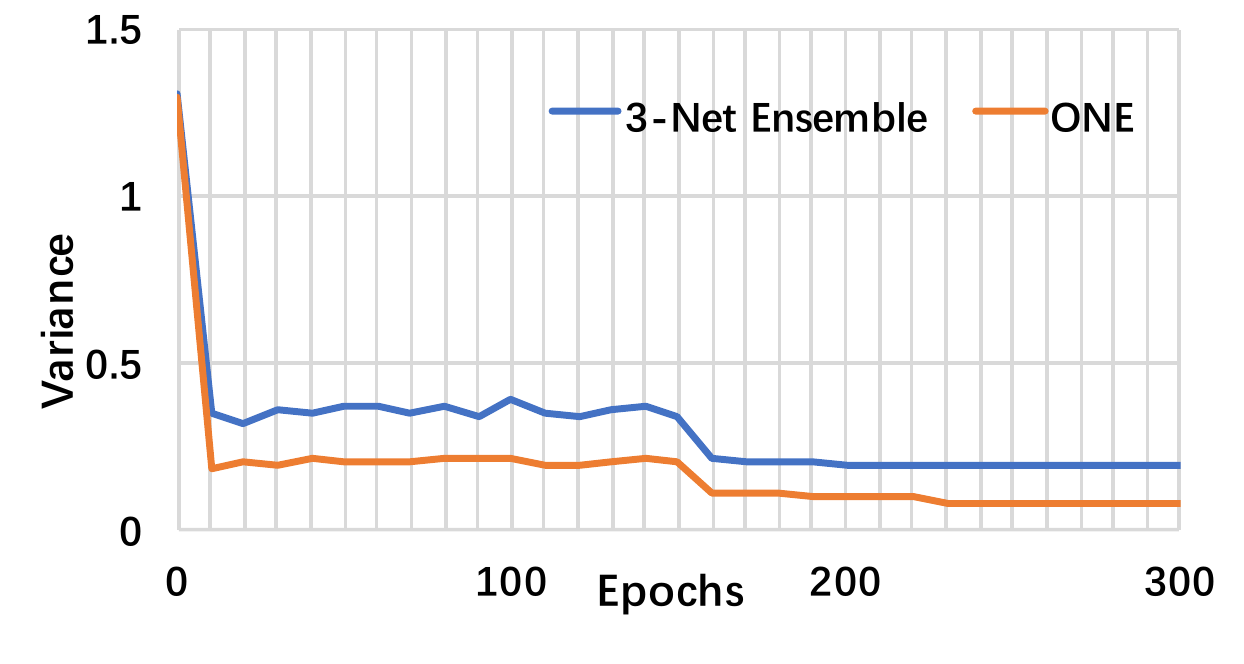}
	\caption{Model variance during training.}\label{fig:Variance} 
\end{wrapfigure}
\subsection{Variance Analysis on ONE's Branches}
We analysed the variance of ONE's branches over the training epochs
in comparison to the conventional ensemble method.
We used ResNet-32 as the base net and test CIFAR100. 
We quantified the model variance 
by the average prediction differences on training samples
between every two models/branches in Euclidean space.
Fig. \ref{fig:Variance} shows that a 3-Net Ensemble involves
{\em larger} inter-model variances than ONE with 3 branches throughout the training process.
This means that the branches of ONE have higher correlations, due to the proposed
learning constraint from the distillation loss that enforces them align to the same teacher prediction,
which probably hurts the ensemble performance.
However, in the mean generalisation capability (another fundamental aspect in ensemble learning), 
ONE's branches (the average error rate 26.61$\pm$0.06\%) are much superior to
individual models of a conventional ensemble (31.07$\pm$0.41\%),
leading to stronger ensembling performance.


\section{Conclusion}

In this work, we presented a novel On-the-fly Native Ensemble (ONE) strategy for 
improving deep network learning through online knowledge distillation
in a one-stage training procedure. 
With ONE, we can more discriminatively learn both
small and large networks with less computational cost,
beyond the conventional offline alternatives that are typically
formulated to learn better small models alone.
Our method is also superior
over existing online counterparts due to the unique capability of
constructing a high-capacity online teacher 
to more effectively mine knowledge from the training data and 
supervise the target network concurrently.
%
Extensive experiments 
show that a variety of deep networks 
can all benefit from the ONE approach 
on four image classification benchmarks.
Significantly, smaller networks obtain more performance gains,
making our method specially good for low-memory and fast execution scenarios.
%

%
%

\section*{Acknowledgements}
{\small This work was partly supported by the China Scholarship Council, Vision Semantics Limited, the Royal Society Newton Advanced Fellowship Programme (NA150459), and Innovate UK Industrial Challenge Project on Developing and Commercialising Intelligent Video Analytics Solutions for Public Safety (98111-571149).}
\bibliographystyle{unsrt}
\bibliography{nips18}

\end{document}